\documentclass[a4paper,USenglish,cleveref, autoref, thm-restate]{oasics-v2021}

\newtheorem{hypothesis}[theorem]{Hypothesis}
\title{An Explainable GNN Framework for Component-Level Anomaly Diagnosis} 

\author{Sena Ozgunay}{LAAS-CNRS, Université de Toulouse, France}  {sozgunay@laas.fr}{https://orcid.org/0009-0009-4991-2215}{}

\author{Louise Travé-Massuyès}{LAAS-CNRS, Université de Toulouse, France} {louise@laas.fr}{https://orcid.org/0000-0002-5322-8418}{}

\author{Jean-Michel Loubes}{INRIA, France}  {jean-michel.a.loubes@inria.fr}{https://orcid.org/0000-0002-1252-2960}{}

\author{Raul Sena Ferreira}{Airbus Protect, Toulouse, France}  {raul.sena-ferreira@airbus.com}{https://orcid.org/0000-0002-4629-8821}{}

\authorrunning{S. Ozgunay, L. Travé-massuyès, JM. Loubes and RS. Ferreira} 

\Copyright{Sena Ozgunay, Louise Travé-Massuyés, Jean-Michel Loubes and Raul Sena Ferreira}



\category{} 

\relatedversion{} 

\supplement{}
\supplementdetails[subcategory={Source Code}, cite={}, swhid={}]{Software}{https://github.com/sozgny/XS2C}

\acknowledgements{This work is supported by the AI Interdisciplinary Institute ANITI, funded by the France 2030 program under the Grant agreements n° ANR-19-PI3A-0004 and n° ANR-23-IACL-0002. The authors would also like to thank Ferhat Tamssaouet for valuable support and insightful discussions for this work.}

\nolinenumbers 

\EventEditors{Ingo Pill, Marina Zanella, and Gregory Provan}
\EventNoEds{3}
\EventLongTitle{37th International Conference on Principles of Diagnosis and Resilient Systems (DX 2026)}
\EventShortTitle{DX 2026}
\EventAcronym{DX}
\EventYear{2026}
\EventDate{September 1--3, 2026}
\EventLocation{Cork, Ireland}
\EventLogo{}
\SeriesVolume{148}
\ArticleNo{3}
\begin{document}
\maketitle
\begin{abstract}
Industrial processes are complex systems composed of multiple interacting sensors that generate multivariate time series (MTS). Detecting anomalies in such systems is critical for reliability and safety, yet understanding their origin is equally important. Existing Graph Neural Network (GNN)-based methods for anomaly detection primarily focus on sensor-level deviations and either attribute anomalies directly to the deviating sensors. When diagnosis is attempted, generally, the most deviated sensor is identified as a root cause of a system fault. However, in many industrial systems, anomalies do not arise from faulty sensors but from disruptions in the influences governing the system dynamics. We propose an explainable GNN-based anomaly detection framework that shifts the perspective from sensor-level anomalies to component-level diagnosis, hypothesizing that anomalous measurements are symptoms of altered inter-sensor influences. Experiments show that the method effectively identifies and prioritizes the true faulty components, providing interpretable insights into system failures. 
\end{abstract}

\section{Introduction}
Detecting anomalies in industrial systems is insufficient for understanding their origins, especially in complex processes with strongly interconnected sensor measurements. Graph Neural Networks (GNNs) capture such dependencies through a graph where nodes are sensors and edges are their interactions~\cite{scarselli2008graph}, and their message-passing mechanism has made them an effective tool widely used for anomaly detection in multivariate time series~\cite{deng2021graph, yuan2025comprehensive}.

However, detection alone leaves the cause of the failure unresolved. In GNN-based frameworks, diagnosis is occasionally addressed by attributing anomalies to the most deviated sensors~\cite{deng2021graph, liu2024graph, jiang2024hierarchical, li2026graph} or by ranking their contributions~\cite{zhao2020multivariate, zheng2023correlation, zhang2024physical, guo2024knowledge}. Yet, in many industrial settings, anomalies do not originate from sensors but from physical components that disrupt the influences governing the system dynamics. Functional sensors may then produce anomalous measurements due to perturbations: for instance, a pipe leak in a water distribution network alters flow dynamics and yields abnormal readings from multiple healthy sensors~\cite{gardharsson2022graph}. This highlights the need to move beyond sensor-level analysis and consider the underlying sensor influences when performing anomaly diagnosis. We therefore argue that anomalous measurements should be interpreted as \emph{symptoms} rather than faults. This perspective is naturally aligned with model-based diagnosis (MBD)~\cite{reiter1987theory, kleer1986reasoning}, where a consistency-based diagnosis identifies a subset of components that, assumed faulty, makes the system description consistent with the observations. We adopt this principle by treating the graph structure as the system description and the inter-sensor influences as the components to be diagnosed, with anomaly-detection results as observations.

In this paper, we propose the XS2C (eXplainable Sensor-to-Component) framework, which integrates an attention-based GNN with consistency-based diagnosis to identify diagnoses for system anomalies interpreted as symptoms of faults. To the best of our knowledge, no prior method has jointly leveraged these two approaches in a unified manner. 
In summary, the contributions of this work are as follows:
\begin{enumerate}
    \item An adaptation of the Graph Deviation Network (GDN)~\cite{deng2021graph} for anomaly detection, including a graph construction strategy based on Neural Granger Causality (NGC) and a sensor-wise anomaly thresholding.
    \item A fine-tuning of the model with test data containing anomalies to learn the abnormal behavior of the system and computing attention changes of influences based on the model's pre- and post-finetuning states.
    \item The identification of diagnoses for the system anomalies using path search in the graph structure and consistency-based diagnosis reasoning. 
    \item The generation of a global explanation consisting of ranked diagnoses based on the attention changes.
\end{enumerate}
The rest of this paper is organized as follows. Section~\ref{related_work} reviews related work. Section~\ref{framework} presents our framework. In Section~\ref{experiments}, we demonstrate the effectiveness of the proposed method. Section~\ref{conclusion} shows our conclusions and future works.

\section{Related work} \label{related_work}
In the real world, MTS data often lack an explicit graph describing the dependencies among variables, yet such a graph is a crucial input for GNNs. This has motivated a growing body of work on graph construction/learning, where the graph is either learned jointly with the predictive model~\cite{deng2021graph, zheng2023correlation} or inferred beforehand, computing similarity metrics~\cite{jin2023spatio}, using separate algorithms~\cite{liu2024graph, yan2024dynamic}, or introducing domain knowledge~\cite{zhang2024physical, gardharsson2022graph}. For instance, GDN~\cite{deng2021graph} learns a graph structure end-to-end using cosine similarity between learnable node embedding vectors, which might be called global node representations. The graph structure is particularly important because anomaly detection and diagnosis are not only about accurate prediction but also about generating meaningful explanations by identifying which graph interactions are responsible for system anomalies. Among these strategies, Neural Granger Causality (NGC)~\cite{tank2021neural} is particularly attractive, as it can capture directed, data-driven influences between sensors. These causal influences are expected to provide more information and greater interpretability for explaining anomalies than other graph construction/learning strategies. 

Numerous GNN-based anomaly detection methods have been proposed for MTS. They generally aim to learn the system's normal behavior and then detect anomalies by analyzing deviations in sensor measurements. This analysis can be achieved using a reconstruction-based approach or a forecasting-based approach. Reconstruction-based approaches aim to rebuild the given input MTS values by accurately reproducing normal inputs but fail to reconstruct abnormal observations~\cite{zhang2022grelen, li2026graph}. 
Forecasting-based methods focus on predicting future MTS values based on past values and detecting anomalies by the deviations from observations~\cite{deng2021graph, chen2023multivariate, zheng2023correlation}.
Indeed, GDN~\cite{deng2021graph} employs an adaptive version of GAT that integrates global and local node representations for attention-based forecasting. For each forecast time tick, GDN computes an anomaly score for each sensor, then combines them to obtain a global anomaly score for that time tick to determine whether it is an anomaly. In the literature, reconstruction and forecasting strategies are combined to detect anomalies based on predictions from reconstruction and forecasting probabilities~\cite{zhao2020multivariate, gardharsson2022graph}. While these models can effectively detect anomalous patterns, their output is usually limited to anomaly scores or flagged time ticks, leaving the origin/root cause of the anomalies unresolved.

The use of attention mechanisms within GNNs, particularly in a Graph Attention Network (GAT)~\cite{velickovic2018graph}, has gained significant importance in recent years, as attention coefficients can catch most relevant sensor influences in the system by assigning importance scores to edges for a target node. However, the interpretability of attention as explanation still remains a debated topic~\cite{jain2019attention, wiegreffe2019attention, bibal2022attention}. The role of attention in anomaly explanation remains controversial, as it may not correspond to a faithful explanation of model predictions. In MTS, however, attention can still provide a useful signal about changes in sensor influences, as an indicator of how an influence evolves between normal and faulty system conditions. 

\section{Proposed Framework}
\label{framework}
\subsection{Overview}
The overall architecture of XS2C is given in Figure~\ref{framework}. We consider a multi-sensor system with $n$ sensors represented as an MTS and assume the following hypotheses:
\begin{hypothesis}\label{hyp:first}
    In normal conditions, sensor measurements are consistent with the underlying system relationships.
\end{hypothesis}

\begin{hypothesis}\label{hyp:second}
Under faulty conditions, sensor functionality remains intact, with abnormal sensor measurements resulting from system faults.
\begin{itemize}
    \item An anomaly in a sensor measurement is a symptom.
    \item Anomalies result from alterations in the relationships between measured quantities. In other words, when dependencies between quantities are altered, some measurements exhibit unexpected behavior.
\end{itemize}
\end{hypothesis}

In the context of semi-supervised learning, our training data over $T_{train}$ time ticks, $X_{train} \in \mathbb{R}^{n \times T_{train}}$, consists solely of normal observations to learn the system's normal behavior. We denote the learned causal graph from $X_{train}$ using NGC as $\hat{G} =(V, E)$ where $V= \{1, \dots , n\}$ is the set of nodes/sensors, and $E \subseteq V \times V$ is the set of edges/influences.
The first objective is to detect anomalies in test data $X_{test}\in \mathbb{R}^{n\times T_{test}}$. At time $t\in T_{test}$, we detect which sensor measurement(s) is/are anomalous, i.e., $\hat{y}_i(t)\in\{0,1\}, i \in V$, with $0$ indicating the normal class and $1$ the anomaly class. If ground truth time tick labels are only available at the system level, the sensor-level status is aggregated as $\hat{y}(t)=1 \text{ if } \exists i \in V \text{ such that } \hat{y}_i(t)=1$. The GNN anomaly detection module is based on an attention mechanism, outputting attention coefficients $\alpha_{ij}(t)$ on the edges from a neighbor node $j\in V$ to a target node $i\in V$, and detecting abnormal nodes.
The second objective is to diagnose anomalies. Here, we fine-tune our base model on $X_{test}$ to learn the system's abnormal behavior. We then compute changes in edge attention to capture the evolution between normal and faulty system states. We compute diagnoses from nodes detected as abnormal via path search in the graph, leveraging the consistency-based diagnosis theory~\cite{reiter1987theory}. The output is a global explanation of the system fault and ranked diagnoses based on attention changes.
\begin{figure}[t]
    \centering
    \includegraphics[width=\linewidth]{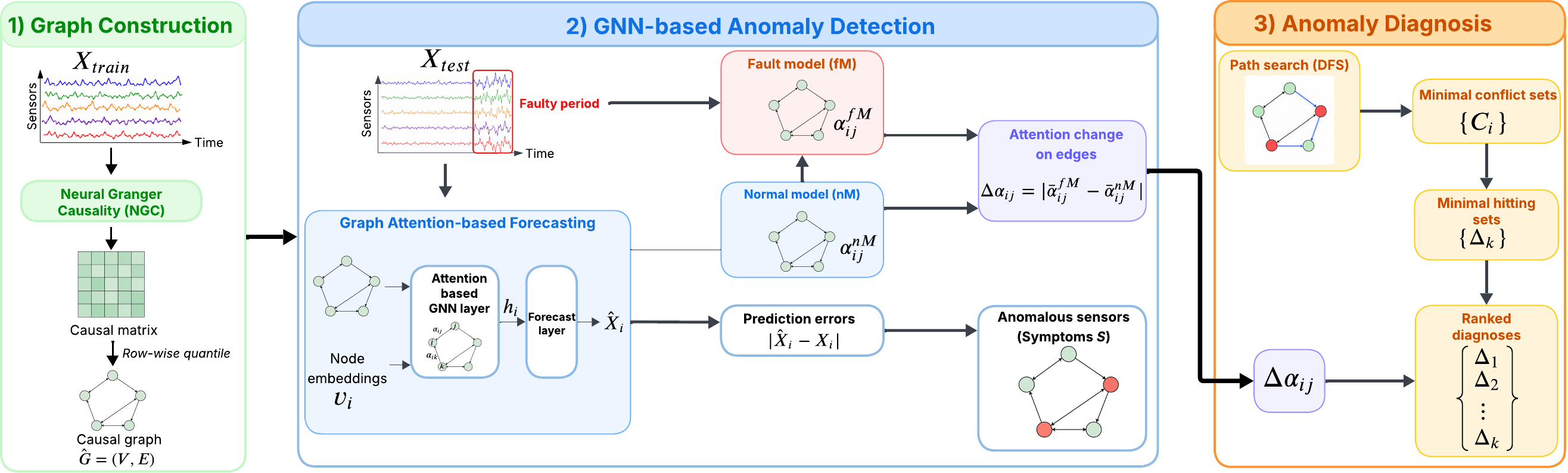}
    \caption{\textbf{Our framework is decomposed into three components: \textit{Graph construction, GNN-based anomaly detection, and anomaly diagnosis}}. In the \textit{graph construction}, a causal graph is extracted from $X_{train}$ using NGC. For the \textit{GNN-based anomaly detection}, an attention-based forecasting is applied on $X_{test}$ based on the graph structure. Symptoms are identified for the next step. During the fault period, we fine-tune the model on $X_{test}$ to capture the system behavior under faulty conditions. Attention changes on edges are computed based on the model's pre- and post- finetuning states. In the \textit{anomaly diagnosis} step, the symptoms and attention changes are then used to generate ranked diagnoses to identify faulty system components.}
    \label{fig:framework}
\end{figure}
In summary, XS2C has three components: \textit{graph construction, GNN-based anomaly detection}, and \textit{anomaly diagnosis}, as detailed in the next subsections.

\subsection{Graph Construction}
In case of no prior graph, our objective is to construct a directed causal graph from $X_{train}$ using NGC~\cite{tank2021neural}. A variable $j\in \{1,\dots,n\}$ Granger-causes a variable $i\in \{1,\dots,n\}$ if knowledge of the past of $j$ improves the prediction of the future of $i$. In NGC, we quantify the temporal influence of $j$ on $i$ by modeling each sensor as a neural network and extracting the Granger causal structure by forcing some sets of weights to zero via group penalties. For a sensor $i$ at time $t$, we define:
\begin{align} \label{ngc_def}
    \hat{x}_i(t)= G_i(x_1(<t), \dots , x_n(<t)) + \epsilon_i(t)
\end{align}
where $x_j(<t)= (x_j(t-K), \dots , x_j(t-2),x_j(t-1))$ are the inputs of $j$ for past $K$ lags, and $\epsilon_i (t)$ is a zero-mean noise. As a neural model $G_i$, we use a component-wise MLP (cMLP)~\cite{tank2021neural}. Only the first layer is used to infer causal relationships, since it directly maps the lagged input variables to the target variable. Hence, the weights of the first layer are divided into blocks by time lags $W^1 = \{W^{11}, \dots, W^{1k}, \dots, W^{1K}\} \in \mathbb{R}^{H \times (nK)}$, where $H$ is the hidden layer dimension and $W^{1k} \in \mathbb{R}^{H \times n}$ captures the contribution of inputs at lag $k$. For a source variable $j$, its overall influence on the target $i$ is determined by the collection $W_j^1 = \{W_{:,j}^{11}, \dots , W_{:,j}^{1K} \}$, where $W_{:,j}^{1k} \in \mathbb{R}^H$ denotes all connections from $j$ to the hidden layer at lag $k$.

A group-lasso penalty is then applied to these weights, such that if all corresponding weights are shrunk to zero, variable $j$ is considered not to Granger-cause variable $i$. We consider this penalty as a causal score $s_{ij} = ||W_j^1||_F$, where $||.||_F$ is the Frobenius matrix norm, and $s_{ij}$ is the causal influence from the source sensor $j$ to the target sensor $i$. We then convert these causal scores into a directed graph by retaining the source sensors whose causal scores exceed a row-wise quantile threshold $q$ as neighbors of $i$: $\mathcal{N}_i =\{ j \ | s_{ij} \geq Q_i(q), j \neq i \}$, where $Q_i(q)$ is the $q$-th quantile of the causal scores in row $i$. Thus, a directed causal graph $\hat{G}=(V, E)$ is obtained from $X_{train}$ with $E = \{ e_{ji}\ | \ j \in \mathcal{N}_i \}$ and used as a fixed structure in the rest of themethodology, ensuring causally meaningful edges and a stable, interpretable system. 

\subsection{GNN-based Anomaly Detection}
\label{sec:gnn_based_ad}
We deploy the attention-based forecasting architecture of the GDN~\cite{deng2021graph} to predict sensor measurements at time $t$ based on a historical sliding window of size $w$:
\begin{align} \label{window}
    [X(t-w),\ X(t-w+1),\ \dots ,\ X(t-1)] \in \mathbb{R}^{n \times w}
\end{align}

\subsubsection{Attention-based Forecasting}
First, each sensor $ i \in \{1,\dots, n\}$ is represented with a global embedding vector that captures its characteristics: $v_i \in \mathbb{R}^d$, where $d$ is the embedding dimension. These vectors are initialized randomly and trained with the model to capture the underlying factors that shape nodes' different behaviors. Given a historical window as in~\ref{window}, the message-passing mechanism computes node $i$'s aggregated local representation via an attention mechanism enhanced with global sensor embeddings, unlike GAT~\cite{velickovic2018graph}. At time $t$, attention coefficients $\alpha_{ij}(t)$ are computed as follows:
\begin{align}\label{attention computation}
    g_i(t)    &= v_i \oplus \operatorname{W} X_i (t)\\
    \phi (i,j) &= \operatorname{LeakyReLU}(\operatorname{a}^\top(g_i(t) \oplus g_j(t))\\
    \alpha_{ij}(t) &= \frac{\operatorname{exp}(\phi(i,j))}{\sum_{k\in \mathcal{N}(i)\cup {i}}\operatorname{exp}(\phi(i,k))}
\end{align}
where $\operatorname{W}\in \mathbb{R}^{d \times w}$ is a trainable weight matrix, $X_i(t) \in \mathbb{R}^w$ is the node $i$'s input measurements, $\operatorname{a}^\top \in \mathbb{R}^{4d}$ is a learned attention weight vector. The node representation is then:

\begin{align}\label{node_repr}
    h_i(t) = \operatorname{ReLU} \left( \alpha_{ii}(t) \operatorname{W}X_i(t) + \sum_{j\in \mathcal{N}_i} \alpha_{ij}(t)\operatorname{W}X_j(t)\right)
\end{align}
After obtaining all final local node representations $h_i(t), i=\{1, \dots,n\}$, we predict the sensor measurements at time $t$ as:
\begin{align}\label{pred}
    \hat{X}(t) = f_{\theta} \left([v_1\circ h_1(t), \dots, v_n \circ h_n(t)]    \right) \in \mathbb{R}^n
\end{align}
where $\circ$ is the element-wise multiplication and $f_{\theta}$ is an MLP. The training process is achieved by minimizing the Mean Squared Error (MSE) loss, which is computed as follows:

\begin{align}\label{loss_fct}
    \mathcal{L}_{MSE} = \frac{1}{T_{train}-w}\sum_{t=w+1}^{T_{train}} ||\hat{X}(t)- X(t)||_2^2
\end{align}

\subsubsection{Computing thresholds and detecting anomalies} 
Traditional MTS anomaly detection methods use a single threshold based on aggregated anomaly scores to detect abnormal time ticks at the system level~\cite{deng2021graph, zhao2020multivariate, zheng2023correlation}. GDN~\cite{deng2021graph}, for instance, computes per-sensor scores but applies a max-based thresholding at the system level, which often ignores gradual or sensor-specific anomalous behaviors in favor of the largest immediate deviations. However, in real-world systems, sensors might measure different quantities, leading to heterogeneity in abnormality. That is why we adopt a sensor-wise thresholding strategy to detect anomalous sensor measurements and subsequently identify them as symptoms. For each sensor, we compute an anomaly score based on its prediction errors at time $t$: $\mathcal{A}_i(t) = | X_i(t) - \hat{X}_i(t)|$. On the validation set $T_{val}$ (containing only normal measurements), we collect $\mathcal{A}_i^{val} = \{\mathcal{A}_i(t) \mid t\in T_{val}\}$ and adaptively select the quantile level matching a target positive prediction rate (PPR) $p_{target}$, which controls the expected false alarm behavior $q^* = \operatorname{arg min}_q |p_{val}(q) - p_{target} |$, where $p_{val}(q)$ is the empirical proportion of time ticks flagged as anomalous in validation data when using quantile $q$. The sensor-wise threshold is then $\tau_i = Q_{\mathcal{A}_i^{val}} (q^*)$. At $t \in T_{test}$, if $\mathcal{A}_i(t) > \tau_i$, we detect $t$ as anomalous, i.e., $\hat{y}_i(t)=1$.

\subsubsection{Identifying symptoms}
Given the binary anomaly predictions for each sensor over $X_{test}$, we compute the \emph{anomaly occurrence} for each sensor, particularly in the faulty period $T_{f} \subseteq T_{test}$. For a sensor $i$, its anomaly occurrence in $T_{f}$ is computed as follows:
\begin{align}
    O_i = \sum_{t\in T_{f}} \hat{y_i}(t)
\end{align}

\emph{Symptoms} are identified on the sensors with the highest anomaly occurrence. Sensors are sorted as $O_{i_1} \geq O_{i_2} \geq \dots \geq O_{i_n}$, and the largest gap between consecutive values $d_k = O_{(k)}-O_{(k+1)}$ defines a cut point: 
\begin{align}
    k^* =\operatorname{argmax}_k d_k
\end{align} 
This criterion identifies the point of maximum separation between highly anomalous sensors and the remaining ones, allowing for a data-driven selection of symptoms: 
\begin{align}
    S =\{ i_1, i_2 , \dots , i_{k^*}\}
\end{align}

\subsection{Anomaly Diagnosis}
We aim to generate a global explanation of the set of detected symptoms in the test data $X_{test}$.
As detailed below, our approach first quantifies the changes of the attention coefficients given by~\eqref{attention computation} associated with the graph's edges from the GNN normal model to the faulty one and then uses these changes to rank the diagnoses obtained via the theory of model-based diagnosis along the consistency-based approach~\cite{reiter1987theory}, which takes the graph of the GNN as a model.

\subsubsection{Fine-tuning of the GNN model}
We consider two models trained under normal and faulty operating conditions:
\begin{itemize}
    \item The \textit{normal model} ($\operatorname{nM}$) is the base GNN model trained on the normal train set $X_{train}$ and tested on $X_{test}$ containing anomalies.
    \item The \textit{fault model} ($\operatorname{fM}$) is a fine-tuned, semi-frozen version of the normal model. 
    The semi-frozen GNN is retrained and tested only on the data collected after a faulty occurrence in $X_{test}$. We assume faults to be strong and persistent, i.e., once a fault occurs, its effects remain present throughout the considered faulty period $T_f$ and continuously affect the system behavior. Consequently, the subsequent occurrence data are assumed to predominantly reflect the faulty operating mode, allowing the fine-tuned model to learn the altered sensor influences associated with the fault.
\end{itemize}

From the normal model, the semi-frozen GNN retains the graph topology ($\hat{G}$) and the global sensor embeddings ($v_i$). The remaining learnable weights are initialized using the weights from the normal model. This strategy allows us to highlight differences between the two models that reflect true behavioral changes—rather than structural variations—particularly in the attention coefficients on the edges. These coefficients capture changes in the interactions between quantities measured by the sensors, providing signals to identify root-cause faulty components in the system.

\subsubsection{Computing attention changes}\label{sec:attention_changes}
The attention coefficients on all forecasted time ticks $t \in T_{f}$ are computed for both models, yielding $\alpha_{ij}^{\operatorname{nM}}(t)$ and $\alpha_{ij}^{\operatorname{fM}}(t)$. Since we aim at a global explanation, we average them over $T_{f}$:
\begin{align} \label{mean_attention}
    \bar{\alpha}_{ij}^{\operatorname{nM}} = \frac{1}{T_{f}}\sum_{t\in T_{f}}\alpha_{ij}^{\operatorname{nM}}(t)\\
    \bar{\alpha}_{ij}^{\operatorname{fM}} = \frac{1}{T_{f}}\sum_{t\in T_{f}}\alpha_{ij}^{\operatorname{fM}}(t)
\end{align}
Then, we compute absolute differences of average attentions,  $\Delta \alpha_{ij} = | \bar{\alpha}_{ij}^{\operatorname{fM}} - \bar{\alpha}_{ij}^{\operatorname{nM}} |$, which will be considered as attention changes of influences.

\subsubsection{Bridging with consistency-based diagnosis} 

The consistency-based diagnosis approach, originally developed using logical models~\cite{reiter1987theory, de1987diagnosing}, has been explored in recent works integrating model-based reasoning with data-driven learning, often through residual classifiers or learned representations~\cite{wilhelm2021overview, jung2019isolation}. 
In this work, we propose using this approach to explain detected anomalies by identifying the sets of faulty components that may have caused them.

Formally, a \emph{system} is defined by the triplet
$(\operatorname{SD} ,\operatorname{COMPS}, \operatorname{OBS})$, 
where $\operatorname{SD}$ is the \emph{system description},
$\operatorname{COMPS}$ is the \emph{set of components}, and 
$\operatorname{OBS}$ is the \emph{set of observations}. A system description describes how the components normally behave by appealing to the distinguished predicate $AB$ whose negation $\neg AB$ has the intended meaning `normal'~\cite{reiter1987theory}. In our setting, this triplet is defined as:

\begin{itemize}
    \item $\operatorname{SD}$ is the graph $\hat{G}$ composed of sensors and influences. 
    \item $\operatorname{COMPS}$ are the influences, since they can be associated with the physical components that create them, i.e., $\operatorname{COMPS} = E$.
    \item $\operatorname{OBS}$ are defined by symptomatic labels $l_i \in \{0,1\}$, where $l_i=1$ indicates that sensor $i$ is a symptom.
\end{itemize}

\begin{definition}[Diagnosis] 
    A diagnosis for the system $(SD, COMPS, OBS)$  is a set  $\mathcal{D} \subseteq COMPS$ such that $SD \cup OBS \cup \{AB(e) \mid e \in \mathcal{D}\} \cup \{ \neg AB(e) \mid e \in COMPS - \mathcal{D}\}$ is satisfiable.
\end{definition}

This implies that the hypothesis—where the influences in $\mathcal{D}$ are considered faulty, and all others are normal—is consistent with both $OBS$ and $SD$. A diagnosis thus assigns a mode (normal or faulty) to each graph influence in a way that aligns with the model and the observations.

\begin{definition}[Minimal diagnosis] 
    A \emph{minimal diagnosis} is a diagnosis $\mathcal{D}$ such that $\forall \mathcal{D}' \subset \mathcal{D}$, $\mathcal{D}'$ is not a diagnosis. 
\end{definition}

Typically, the set of diagnoses is obtained through a two-step process, with the first step grounded in the concept of conflict—originally introduced in~\cite{reiter1987theory} and expanded in~\cite{de1992characterizing}.

\begin{definition}[Conflict and minimal conflict] 
    A \emph{conflict} is a set $\mathcal{C} \subseteq COMPS$ in which all the components of $\mathcal{C}$ are normal is not consistent with $SD$ and $OBS$. A \emph{minimal conflict} is a conflict that does not contain other conflicts.
\end{definition}

As shown by~\cite{reiter1987theory}, minimal diagnoses correspond to the hitting sets~\footnote{A set that intersects every set in the collection.} of the set of minimal conflicts, and the parsimony principle suggests favoring minimal diagnoses. In our setting, the first step for identifying them is to search $\hat{G}$ for the paths $\pi$ that end in a symptom node, start in a non-symptom node, and have no cycles:
\begin{align}
    \begin{cases}
    \pi = (i_k, i_{k-1}, \dots , i_1, i_0),\\
    \forall j \text{ s.t. } 0 \le j \le k-1, i_j \in S, \\
    i_k \notin S, \\
    i_j \neq i_l, j\neq l.
    \end{cases}
\end{align}

In this search, starting from the symptoms, we explore the graph by iteratively visiting its neighboring nodes, stopping once a non-symptom node is reached. This stopping criterion is motivated by the assumption that a faulty influence should manifest through observable deviations, i.e., symptoms: an influence between two non-symptom nodes is unlikely to be the root cause of the system fault, as it would not produce any detectable effect. This reasoning aligns with an exoneration principle~\cite{cordier2004conflicts}: components that do not contribute to any abnormal behavior are excluded from the set of root causes.

For path search, we use a depth-first search (DFS) algorithm. At each step, the state is the triplet $(i_l,\; V_{\text{vis}},\; E_{\text{path}})$, where $i_l \in V$ is the current node, $V_{\text{vis}} \subseteq V$ the set of nodes already visited on the current path, and $E_{\text{path}} \subseteq E$ the set of edges traversed up to $i_l$.

\emph{Transition Rules:} For any incoming edge $e_{l,l-1}=(i_l,i_{l-1}) \in E$, we apply the following rules:
\begin{itemize}
    \item If \( i_l\notin S \), the path is closed and a path is generated $\pi = E_{\text{path}} \cup \{e_{l,l-1}\}$;
    \item If \( i_l \in S \) and \( i_l \notin V_{\text{vis}} \), the procedure continues recursively from \( i_l \);
    \item Otherwise, the transition is ignored to avoid any repetition of the node or cycle.
\end{itemize}

For each $i \in S$, we denote the \emph{set of edge paths}: $\pi_i = \left\{ \pi \;\middle|\; \pi \text{ is an admissible path} \right\}$. The conflict associated with symptom $i$ is defined as the union of the influences appearing in these paths: \begin{equation} 
    \mathcal{C}_i = \bigcup_{\pi\in\pi_i} E(\pi), \label{eq:xs2c_symptom_conflict}
\end{equation} 
where $E(\pi)$ denotes the set of edges belonging to path $\pi$. Since anomalies may propagate through reciprocal interactions between process variables, a faulty bidirectional influence relation is considered capable of covering the conflicts in which it appears, regardless of edge direction. Consequently, reciprocal edges are treated as a single faulty interaction during diagnosis generation; only the direction exhibiting the largest attention change is reported for representation purposes. 

After obtaining all conflict sets $\{C_i \ | \ i \in S \}$, we retain the minimal ones. We then find the minimal hitting set(s) using the \emph{Hitman} solver based on a MaxSAT formulation~\cite{moreno2013implicit} via the PySAT toolkit~\cite{imms-sat18}. Each conflict set $C_i$ is translated into a hard clause $(e_{k,k-1} \vee ... \vee e_{l,l-1} \vee \dots \vee e_{1,0})$ stating that at least one edge must be chosen in each $C_i$, and each unique edge induces a soft clause $\neg e_{l,l-1}$ penalizing its inclusion. These minimal hitting sets correspond to \textit{minimal diagnoses}~\cite{cordier2004conflicts}.

As a global explanation of system anomalies, we aim to rank these diagnoses using attention changes. The attention changes of influences are computed as $\Delta \alpha_{ij}$ in Section~\ref{sec:attention_changes}. If a diagnosis consists of a single influence, its ranking score is directly given by this value. For a diagnosis $\mathcal{D}$ containing multiple influences, we compute the ranking score as the average of the attention changes of its constituent influences:
\begin{align}
    score(\mathcal{D}) = \frac{1}{|\mathcal{D}|}\sum_{e_{ji}\in \mathcal{D}}\Delta\alpha_{ij}
\end{align}
Diagnoses are then ranked by ranking scores in descending order. The intuition is that fault-related propagation should induce larger changes in the corresponding sensor influences, so diagnoses with higher average attention changes are considered more plausible explanations of the observed anomalous behavior. 


\section{Experiments}
\label{experiments}
In this section, we conduct experiments to evaluate the XS2C's anomaly detection performance and the capacity of its anomaly diagnosis component.

\subsection{Experimental Setup}
\subsubsection{Datasets}
\begin{figure}[t]
    \centering
    \includegraphics[width=\linewidth]{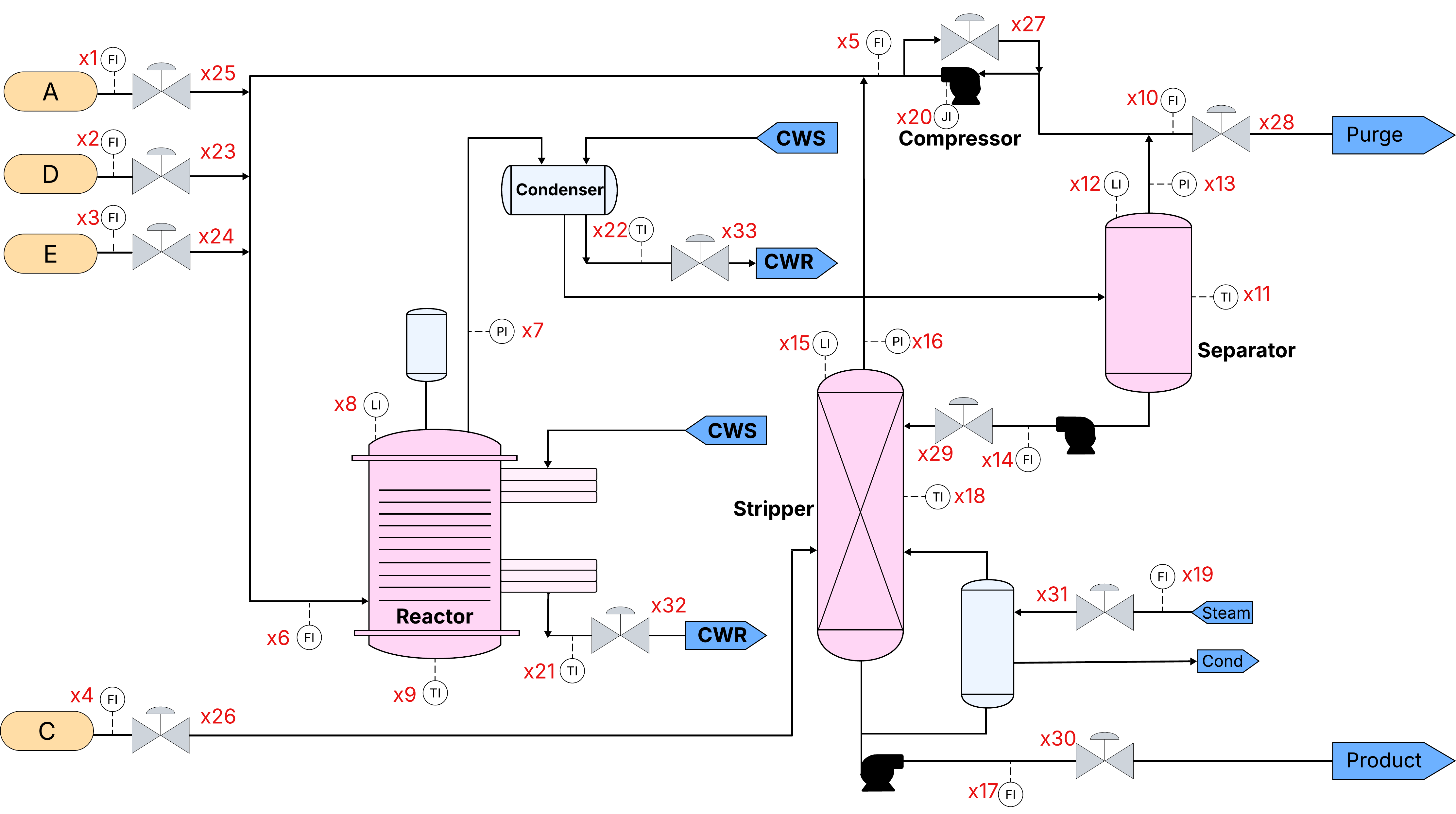}
    \caption{\textbf{TEP flowchart}. The considered process variables are in red text.}
    \label{fig:tep_flowchart}
\end{figure}

We evaluate XS2C on two widely used benchmarks for MTS anomaly detection: TEP and SWaT. More details about the TEP and SWaT can be found in~\cite{downs1993plant, goh2016dataset}.
\begin{itemize}
    \item \textbf{Tennessee Eastman Process (TEP)} is an industrial chemical process for producing two liquid components (G and H) from four gaseous reactants (A, C, D, and E). It consists of five main unit operations (reactor, condenser, liquid-vapor separator, recycling compressor, and product stripper). TEP contains 41 process measurements and 12 manipulated variables. We consider the first 22 processes and 11 manipulated variables, as the remaining measurements are not continuously monitored, and the last variable is constant~\cite{yan2024dynamic}. The TEP variables used in this study are shown in Table~\ref{tab:tep_variables}, and the process is illustrated in Figure~\ref{fig:tep_flowchart}.

    \item \textbf{The  Secure Water Treatment (SWaT)} dataset is a scaled-down industrial water treatment testbed developed by the Singapore Public Utilities Board~\cite{mathur2016swat}. It includes 7 days of normal operation for training and 4 days of attack scenarios for testing, where anomaly labels correspond to injected attacks at different time intervals. . 
\end{itemize}

We use a Python-based simulator\footnote{\url{https://github.com/jkitchin/tennessee-eastman-profbraatz}} for TEP data generation~\cite{book_fd}, operating in \textit{closed-loop} control mode. We simulate 7 days of normal operation for training. For evaluation, 20 different fault scenarios IDV(1-20) are simulated over 48 hours, where the first 8 hours correspond to normal operation followed by the fault period, except for IDV6 and IDV18, whose simulations were terminated after $\thicksim7$ and $\thicksim12$ hours of the faulty period, respectively, due to the violation of the shutdown limits. The simulation yields 20 test datasets with measurements recorded every 3 minutes.  

Following prior works~\cite{deng2021graph, zheng2023correlation}, the first 21,600 samples are removed from the SWaT dataset, and the data is down-sampled to one measurement every 10 seconds using median values. Table~\ref{tab_statistics} reports the statistics of datasets. 

\begin{table}[ht!]
    \centering
    \setlength\extrarowheight{-2pt}
    \resizebox{0.98\textwidth}{!}{
    \begin{tabular}{ll|ll}
        \hline
        \textbf{Variable} & \textbf{Description} & \textbf{Variable} & \textbf{Description}\\
        \hline
         x1& A feed & x18 & Stripper temperature\\
         x2& D feed & x19 & Stripper steam flow\\
         x3& E feed & x20 & Compressor work\\
         x4& A and C feed & x21 & Reactor cooling water outlet temperature\\
         x5& Recycle flow & x22 & Separator cooling water outlet temperature\\
         x6& Reactor feed rate & x23 & D feed flow\\
         x7& Reactor pressure & x24 & E feed flow\\
         x8& Reactor level & x25 & A feed flow\\
         x9& Reactor temperature & x26 & A and C feed flow\\
         x10& Purge rate & x27 & Compressor recycle valve\\
         x11& Product separator temperature& x28 & Purge valve\\
         x12& Product separator level& x29  & Separator pot liquid flow\\
         x13& Product separator pressure& x30 & Stripper liquid product flow\\
         x14& Product separator underflow& x31 & Stripper steam valve\\
         x15& Stripper level& x32 & Reactor cooling water flow\\
         x16& Stripper pressure& x33& Condenser cooling water flow\\
         x17& Stripper underflow &  & \\
         \hline
    \end{tabular}}
    \caption{TEP variables used in this study.}
    \label{tab:tep_variables}
\end{table}

\begin{table}[ht!]
    \centering
    \small
    \caption{Statistics of datasets.}
    \resizebox{0.8\textwidth}{!}{
    \begin{tabular}{lccccc}
        \hline
        \textbf{Datasets} & \textbf{\#Variables} & \textbf{\#Train} & \textbf{\#Test} & \textbf{Anomalies} \\
        \hline
        TEP (IDV1-5,7-17,19-20)  & 33  & 3330       & 930   & 86\% \\
        TEP (IDV6) & 33  & 3330       & 254   & 51.18\% \\
        TEP (IDV18) & 33  &   3330     & 563   & 23.09\% \\
        SWaT & 51  &  47515 & 44986 & 11.97\% \\
        
        \hline
    \end{tabular} }
    \label{tab_statistics}
\end{table}

\subsubsection{Baselines} We compare our anomaly detection component with three representative GNN-based frameworks for MTS anomaly detection: GDN~\cite{deng2021graph}, MTAD-GAT~\cite{zhao2020multivariate}, and CST-GL~\cite{zheng2023correlation}.
\begin{itemize}
    \item \textbf{GDN} learns a graph structure end-to-end using cosine similarity between learnable node embedding vectors and applies a GAT-based forecasting as explained in Section~\ref{sec:gnn_based_ad}. For anomaly detection, GDN computes a per-sensor anomaly score at each time tick and then aggregates them into a system-level score through max-based thresholding. 
    \item \textbf{MTAD-GAT} is a hybrid forecasting-reconstruction anomaly detection method that uses two GAT layers; one for the feature dimension and one for the temporal dimension to capture both feature-wise and temporal dependencies. It detects anomalous time ticks using a joint score derived from predictions and reconstruction probabilities.
    \item \textbf{CST-GL} is a forecasting-based anomaly detection method that explicitly captures multivariate correlations through graph learning. It combines graph-convolution and temporal-convolution layers to model spatial-temporal dependencies and identifies time tick anomalies based on forecasting deviations.
\end{itemize}

\subsubsection{Parameter settings}
For the NGC graph learning module, hyperparameters are optimized via random search on $X_{train}$: a cMLP model is trained with the Adam optimizer, combining group-lasso regularization for sparse causal structure learning and ridge regularization for weight stabilization (learning rate and group-lasso coefficient ($\lambda$) are $0.001$)~\cite{tank2021neural}. Configurations are evaluated based on the average MSE loss across all variables. 

The GNN module is trained for up to 100 epochs using the Adam optimizer (lr=0.001), with early stopping patience of 15 and a validation ratio of 0.1. A self-loop is added to each node to enhance the attention-based message passing mechanism. Hyperparameters are selected by empirical search minimizing the validation MSE; the final settings for both stages are given in Table~\ref{tab:params} (NGC: lag $l$, hidden layer dimension $H$, ridge coefficient $\lambda_{ridge}$, quantile $q$ used for causal graph extraction; GNN: batch size $b$, sliding window $w$, node embedding dimension $d$, forecasting layer hidden dimension h, and target positive prediction rate $p_{target}$).

Baselines reproduced from corresponding github repositories are used \footnote{\url{https://github.com/d-ailin/GDN},\url{https://github.com/huankoh/CST-GL}, \url{https://github.com/ML4ITS/mtad-gat-pytorch}} with sliding window size w=5 (30) for SWaT (TEP). On SWaT, the default settings reported by GDN and CST-GL are kept; on TEP, an empirical search is performed around their default values. The main settings are reported as SWaT (TEP): \textbf{GDN} -- 50 epochs (early-stopping patience of 10), batch size 16, embedding dimension 64, forecasting layer hidden dimension 128 (256), neighbor size 15 (3). \textbf{CST-GL} -- 20 epochs, batch size 64, embedding dimension 256, retain ratio 0.1 (0.3), saturation rate 20 (10), neighbor size 15 (30), 2 (3) temporal-convolution layers with 16 (64) channels, 2 graph-convolution layers with 16 channels, 32 skip channels and 64 (128) end channels. \textbf{MTAD-GAT} -- 50 epochs (with dropout 0.3), batch size 64, initial convolution kernel size 7, one GRU layer (150 units), three forecasting fully-connected layers (150 units), and one reconstruction layer (128 units).

\begin{table}[ht!]
\centering
\caption{Parameter settings for the NGC and GNN modules in XS2C.}
\label{tab:params}
\setlength{\tabcolsep}{2.5pt}
\resizebox{0.6\textwidth}{!}{
\begin{tabular}{c|cccc|ccccc}
\hline
\multirow{2}{*}{Dataset} &
\multicolumn{4}{c|}{NGC-Graph Learning} &
\multicolumn{5}{c}{GNN Module in XS2C} \\

\cline{2-10}

&
$l$ &
$H$ &
$\lambda_{ridge}$ &
$q$ &
$b$ &
w &
d &
h &
$p_{target}$ \\

\hline

TEP &
16 & 64  & 0.001  & 0.90 &
16 & 30 & 64  & 256 & 0.09  \\

SWaT &
4 & 128  & 0.0001 & 0.95 &
16 & 5 & 64  & 64 & 0.06  \\


\hline
\end{tabular}
}
\end{table}

\subsubsection{Evaluation metrics}
To evaluate the performance of the GNN anomaly detection model, since only time-tick-level ground truth labels are available, we aggregate sensor-level predictions to obtain a system-level decision: $t$ is classified as anomalous if at least $N \leq n$ sensors are simultaneously flagged as anomalous: $ \hat{y}(t) =1 \ \text{if} \ \sum_{i=1}^n \hat{y}_i(t) \geq N$. This rule reduces spurious detections caused by sensor noise and reflects that a single underlying fault can manifest across multiple sensors. We use $N=2$ for TEP, since faults are reflected by localized deviations affecting a limited subset of process variables, and $N=15$ for SWaT, since attacks typically cause process-wide consequences (tank overflows, underflows, shutdown sequences, flow disruptions affecting multiple stages), so a larger aggregation reduces false alarms from isolated deviations. 

Detection performance is evaluated with the macro-F1 score ($F_1$) and the Matthews Correlation Coefficient (MCC). Macro-F1 ensures a balanced evaluation of normal and anomalous classes; MCC provides a robust global assessment by simultaneously considering true and false positives and negatives, making it more reliable than accuracy across datasets with varying class distributions.

\subsection{Anomaly detection performance}
Table~\ref{tab:detection_results} reports anomaly detection results on TEP (across 20 fault scenarios) and SWaT. On TEP, the anomaly detection component of our proposed framework XS2C has the best overall performance compared to the baseline. Particularly, we achieve substantially improved performance across most fault scenarios compared with the GDN. Moreover, we are competitive with MTAD-GAT and CST-GL by achieving the best performance on several scenarios, particularly those associated with persistent or structured abnormal dynamics such as IDV(4, 10, 12, 13, 16, 17, 18, 19, 20). Lower performance is consistently observed across all methods on disturbances such as IDV (3, 5, 9, 15), known to involve weaker deviations without abnormal dynamics~\cite{liu2024graph, jiang2024hierarchical, guo2024knowledge}. 

On SWaT, we remain competitive with the baseline with a slightly better performance in both $F_1$ and MCC, demonstrating that our proposed framework can handle large industrial cyber-physical systems while maintaining explainability. Despite anomaly explanation and diagnosis being our primary objective of the proposed framework, we demonstrate an effective anomaly detection performance on two well-known benchmarks. Scenarios with high $F_1$ and MCC should be retained for the next step in anomaly diagnosis, since reliable detection is necessary to provide sufficient symptom information. 

\begin{table}[!ht]
\centering
\caption{Anomaly detection results. The best performing method in each experiment is in bold.}
\label{tab:detection_results}
\resizebox{0.9\textwidth}{!}{
    \begin{tabular}{c|cc|cc|cc|cc}
    \hline
    \multirow{2}{*}{Datasets} &
    \multicolumn{2}{c|}{GDN} &
    \multicolumn{2}{c|}{MTAD-GAT} &
    \multicolumn{2}{c|}{CST-GL} &
    \multicolumn{2}{c}{Our AD component} \\
    
    \cline{2-9}
    
    &
    $F_1$ &
    MCC &
    $F_1$ &
    MCC &
    $F_1$ &
    MCC &
    $F_1$ &
    MCC \\
    \hline
    
    IDV1
    & 0.28 & 0.16 
    & \textbf{0.99} & \textbf{0.98} 
    & \textbf{0.99} & \textbf{0.98} 
    & 0.98  & 0.96\\
    
    IDV2 
    & 0.27 & 0.12
    & \textbf{0.96} & \textbf{0.92 }
    & 0.93 & 0.87 
    & 0.95 & 0.90\\
    
    IDV3 
    & \textbf{0.17} & \textbf{0.05} 
    & 0.12 & -0.03 
    & \textbf{0.17} & \textbf{0.05} 
    & 0.15 & 0.02 \\
    
    IDV4 
    & 0.17 & 0.06 
    & 0.26 & 0.14 
    & 0.23 & 0.11 
    & \textbf{0.98} & \textbf{0.97}\\
    
    IDV5 
    & 0.34 & \textbf{0.21} 
    & 0.34 & 0.19 
    & \textbf{0.37} & \textbf{0.21}
    & 0.35 & 0.19 \\
    
    IDV6 
    & 0.75 & 0.60 
    & \textbf{1} & \textbf{0.99 }
    & 0.99 & 0.98 
    & 0.99 & 0.98\\
    
    IDV7 
    & 0.35 & 0.21 
    & \textbf{0.87} & \textbf{0.77} 
    & 0.62 & 0.43 
    & 0.48 & 0.30\\
    
    IDV8 
    & 0.47 & 0.30 
    & \textbf{0.93} & \textbf{0.86} 
    & 0.88 & 0.79 
    & 0.92 & 0.85  \\
    
    IDV9 
    & \textbf{0.16} & \textbf{0.05} 
    & 0.12 & -0.03 
    & \textbf{0.16} & 0.04 
    & \textbf{0.16} & 0.02\\
    
    IDV10 
    & 0.20 & 0.10 
    & 0.18 & 0.08 
    & 0.46 & 0.28 
    & \textbf{0.74} & \textbf{0.58}\\
    
    IDV11 
    & 0.19  & 0.08 
    & 0.40 & 0.24 
    & 0.43 & 0.26 
    & \textbf{0.57} & \textbf{0.38}\\
    
    IDV12 
    & 0.46 & 0.29 
    & 0.90 & 0.82 
    & 0.93 & 0.86 
    & \textbf{0.97} & \textbf{0.94}  \\
    
    IDV13 
    & 0.36 & 0.22 
    & 0.80 & 0.65 
    & 0.76 & 0.61 
    & \textbf{0.84} &\textbf{ 0.72} \\
    
    IDV14 
    & 0.16 & 0.03 
    & 0.99 & 0.98 
    & \textbf{1} & \textbf{0.99 }
    & 0.96 & 0.92 \\
    
    IDV15 
    & 0.17 & \textbf{0.07} 
    & 0.13 & -0.02 
    & \textbf{0.18} & \textbf{0.07 }
    & 0.15 & 0.02 \\
    
    IDV16
    & 0.20 & 0.09 
    & 0.14 & 0.02 
    & 0.48 & 0.30 
    & \textbf{0.71} & \textbf{0.54} \\
    
    IDV17
    & 0.41 & 0.24 
    & 0.77 & 0.62 
    & 0.87 & 0.76 
    & \textbf{0.90} & \textbf{0.82} \\
    
    IDV18 
    & 0.60 & 0.42 
    & 0.91 & 0.83 
    & 0.90 & 0.81
    & \textbf{0.94} & \textbf{0.88} \\
    
    IDV19 
    & 0.17 & 0.06 
    & 0.16 & 0.06 
    & 0.46 & 0.28 
    & \textbf{0.54} & \textbf{0.35 }\\
    
    IDV20 
    & 0.27 & 0.14 
    & 0.49 & 0.31 
    & 0.64 & 0.45 
    & \textbf{0.71} & \textbf{0.54} \\
    \textit{Average}
    & \textit{0.31} & \textit{0.18}  
    & \textit{0.57} & \textit{0.47} 
    & \textit{0.62} & \textit{0.51} 
    & \textit{\textbf{0.70}} & \textit{\textbf{0.59}}\\
    \hline
    SWaT  &
    0.87 & 0.76 & 
    0.39 & 0.17 & 
    0.87 & 0.76 & 
    \textbf{0.88} & \textbf{0.78} \\
    \hline
    \end{tabular}
    }
\end{table}

\subsection{Anomaly diagnosis results and analysis}
\begin{figure}[t] 
    \centering
    \subfloat[IDV1]{\includegraphics[width=0.5\linewidth]{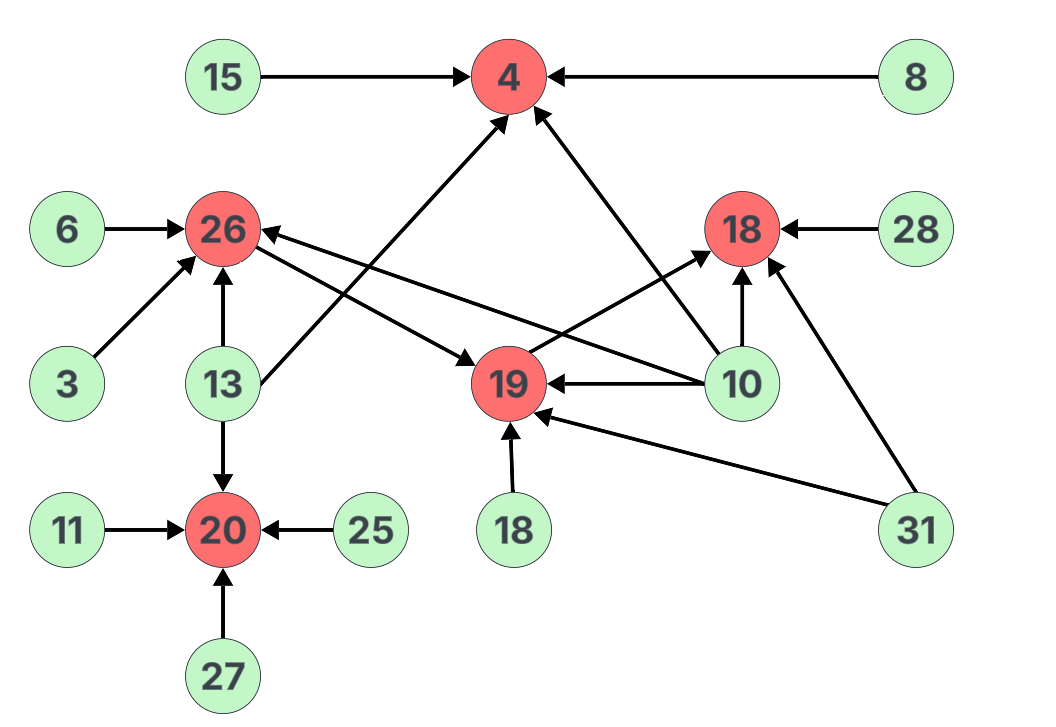}}
    \hfill
    \subfloat[IDV14]{\includegraphics[width=0.3\linewidth]{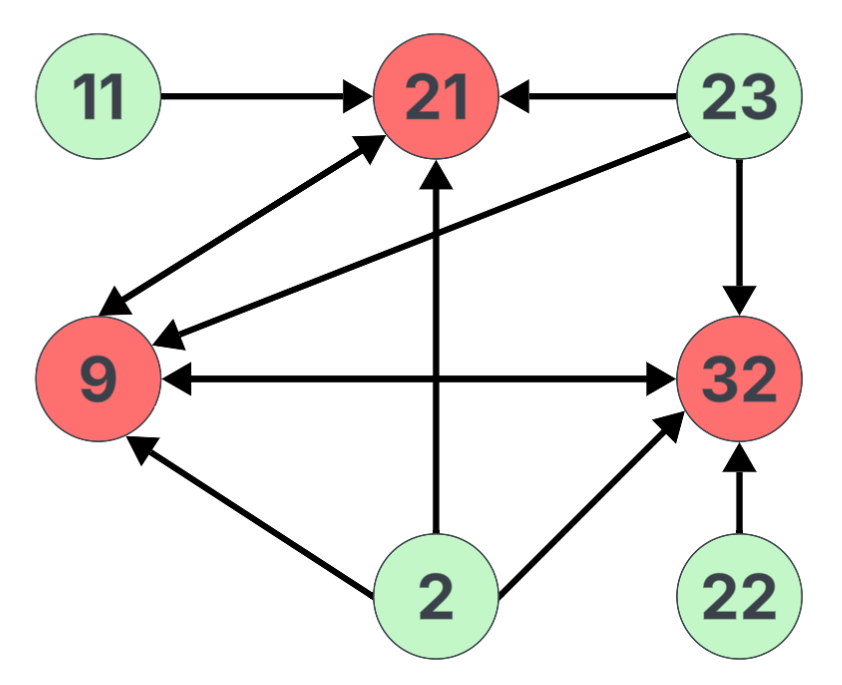}}
    \caption{\textbf{Causal subgraphs obtained for IDV1 and IDV14}: A local subgraph is used for identifying diagnoses regarding neighbors. Red nodes correspond to symptoms, and green nodes are non-symptoms.}
    \label{fig:subgraphs}
\end{figure}

In TEP, IDV1 and IDV14 are chosen as case studies for anomaly diagnosis. For each symptom, a local subgraph is extracted from its neighborhood; these subgraphs serve as the basis for searching paths that yield minimal conflicts and, subsequently, minimal diagnoses. Figure~\ref{fig:subgraphs} illustrates local subgraphs for IDV1 and IDV14, respectively.

IDV1 is a step change disturbance in the A/C feed ratio in stream 4. The identified symptoms are $x_4$ (A and C feed in stream-4), $x_{18}$ (stripper temperature), $x_{19}$ (stripper steam flow), $x_{20}$ (compressor work), and $x_{26}$ (manipulated variable related to A and C feed flow in stream-4). The top-3 ranked diagnoses (with attention changes of edges in them) are:
\begin{align}
    \mathcal{D}_1 &= \{e_{13,4} \ (0.82), e_{25,20} \ (0.04), e_{13,26} \ (0.14)\}\\
    \mathcal{D}_2 &= \{e_{13,4}, e_{13,20} \ (0.02), e_{13,26}\}\\
    \mathcal{D}_3 &= \{e_{13,4}, e_{27,20} \ (0.01), e_{13,26}\}
\end{align}

The nodes involved in the edges composing the diagnoses also include $x_{13}$, the product separator pressure, $x_{25}$, the A feed flow in stream-1, and $x_{27}$, the compressor recycle valve. The diagnoses highlight influences associated with the stream-4 feed mechanism and its downstream propagation, which is coherent IDV1. In the literature, this disturbance is often interpreted through the control system response, where the A feed flow ($x_{25}$) is adjusted to compensate for changes in the feed ratio and is identified as a root-cause variable~\cite{liu2024graph}. Although $x_{25}$ is not a symptom in our framework, it still appears in the top-ranked diagnoses through its perturbed influence, illustrating our hypothesis: the objective is not to identify the symptoms themselves as root causes but the altered influences that explain them. The result should thus not be read as identifying $x_{25}$ itself as an isolated root cause, but as indicating that the dependency involving this variable is significantly perturbed during the fault, consistent with the expected control response to a disturbance in stream-4. Moreover, the repeated involvement of relations connected to $x_{13}$ further suggests that the disturbance propagates downstream through the recycle stream until a new steady state is reached, which agrees with the reported behavior of IDV1. Thus, the proposed diagnosis provides a global explanation: it localizes the disturbance around stream-4 feed interactions while capturing its downstream propagation.

Regarding IDV14, the identified symptoms are the reactors $x_9$ (temperature), $x_{21}$ (cooling water temperature), and $x_{32}$ (cooling water flow). The top-3 diagnoses are:
\begin{align}
    \mathcal{D}_1 &= \{e_{32,9}\ (0.34)\}\\
    \mathcal{D}_2 &= \{e_{2,9} \ (0.15) \}, \{e_{11,21}\ (0.15)\} \\
    \mathcal{D}_3 &= \{e_{23,9}\ (0.14)\}
\end{align}
The nodes involved in the edges composing the diagnoses include $x_{11}$, the product separator temperature, and $x_{23}$, the D feed flow in stream-2. The highest-ranked diagnosis $e_{32,9}$ directly links the reactor cooling water flow $x_{32}$ to the reactor temperature $x_9$. Since the cooling water mechanism directly regulates the reactor temperature, the perturbation of this influence strongly suggests that the fault originates from the reactor cooling water valve system, consistent with previous studies~\cite{liu2024graph, li2026graph}. Compared to approaches that focus only on abnormal variables, our framework provides a more relational interpretation by explicitly identifying the perturbed dependency underlying the cooling process. The remaining diagnoses remain physically meaningful: a sticking valve is expected to alter several temperature-related influences around the reactor~\cite{liu2024graph}, so while $e_{32,9}$ provides the most direct localization, the additional diagnoses support the propagation of the disturbance through the reactor cooling subsystem.

\begin{figure}[t]
    \centering
    \includegraphics[width=0.7\linewidth]{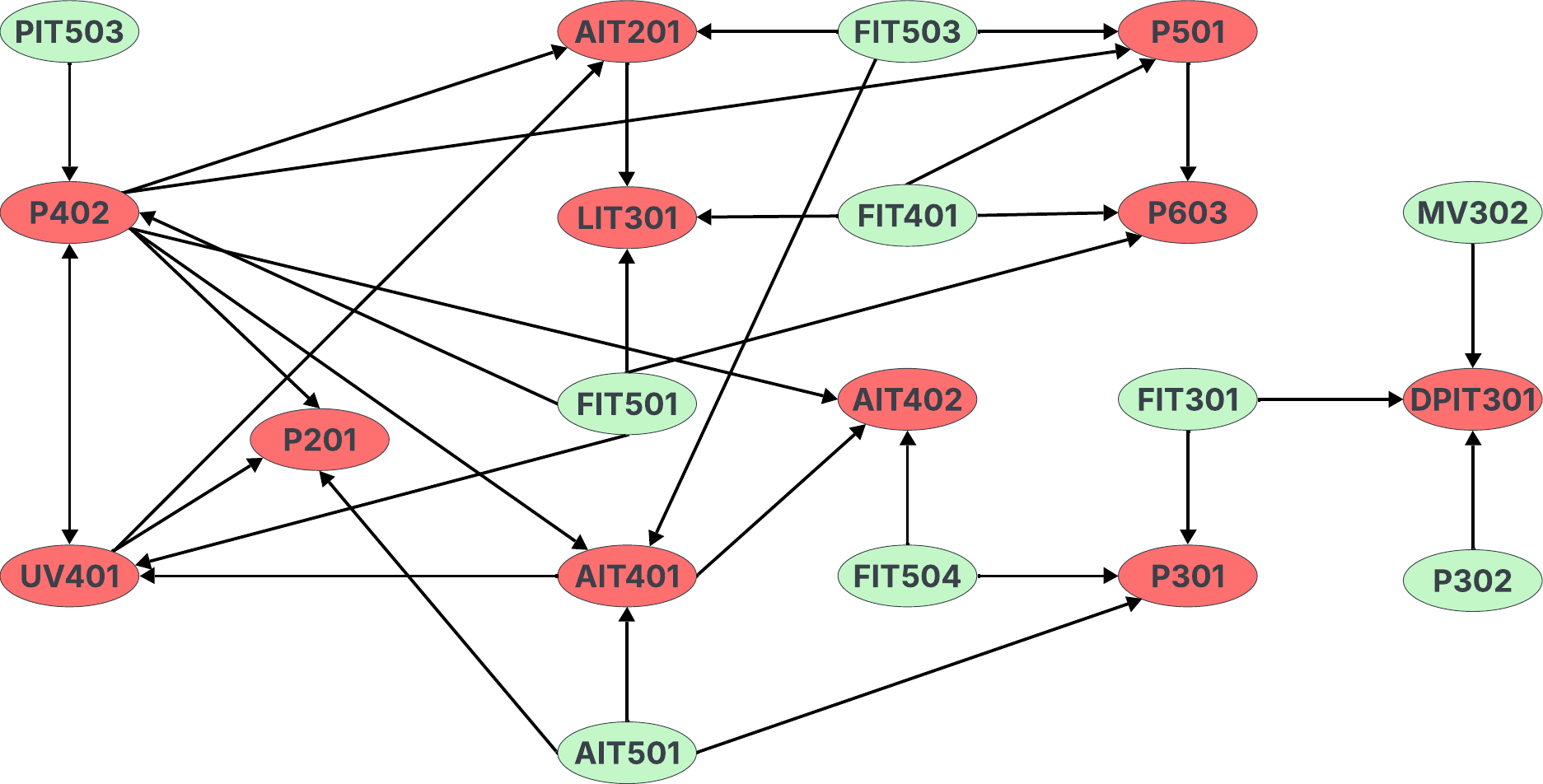}
    \caption{Causal subgraph obtained for SWaT dataset.}
    \label{fig:swat_subgraph}
\end{figure}

For the SWaT case study, we selected the attack scenario corresponding to the continuous closure of valve $MV303$ between 29/12/2015 14:38:12-14:50:08, which disrupts the backwash process in stage-3 (filtering undesirable materials from the water). The identified symptoms mainly involve variables associated with stage-3 and its downstream propagation. Figure~\ref{fig:swat_subgraph} illustrates local subgraph for SWaT. The top-3 diagnoses are:
\begin{equation}
\begin{aligned}
\mathcal{D}_1 = \{&
e_{P302,DPIT301}\,(0.32),\;
e_{AIT501,P301}\,(0.11),\\
&
e_{FIT501,UV401}\,(0.10)
\},\\
\mathcal{D}_2 = \{&
e_{P302,DPIT301}\,(0.32),\;
e_{AIT501,P301}\,(0.11),\\
&
e_{AIT501,AIT401}\,(0.09)
\},\\
\mathcal{D}_3 = \{&
e_{P302,DPIT301}\,(0.32),\;
e_{AIT501,P301}\,(0.11),\\
&
e_{PIT503,P402}\,(0.06)
\}.
\end{aligned}
\end{equation}
In these diagnoses, the influence $e_{P302, DPIT301}$ between the ultra filtration feed pump $P302$ and the differential pressure transmitter $DPIT301$ of stage-3 systematically appears with the largest attention variation, indicating that the hydraulic and backwash dynamics in this region are considerably perturbed. This is physically coherent with a valve-induced disturbance propagation. The other influences appearing in $\mathcal{D}_1$ may be interpreted as secondary propagation effects: $e_{AIT501, P301}$ (water quality analyzer $AIT501$ to pump $P301$) suggests propagation toward the downstream circulation mechanisms, while $e_{FIT501, UV401}$ (flow transmitter $FIT501$ to dechlorinator $UV401$) indicates that abnormal flow conditions induced by the disrupted backwash further impact the UV treatment stage. The diagnoses thus not only localize the disturbance around the backwash drain valve but also reveal plausible downstream propagation pathways.

\section{Conclusion}
\label{conclusion}
This work proposes an explainable anomaly detection framework that shifts diagnosis from anomalous sensors toward disrupted sensor influences. By combining an attention-based GNN with consistency-based diagnosis, our approach, XS2C, produces physically coherent and interpretable diagnoses that localize the disturbed process components/regions underlying system faults. Experimental results demonstrate that XS2C can achieve effective anomaly detection and meaningful explanation of fault propagation dynamics in complex systems. Future work will focus on improving the performance of the anomaly detection component, validating XS2C on real-world industrial systems. Additionally, weakening the Hypothesis \ref{hyp:second} by including the possibility that the sensor might be faulty via self-attention coefficients will be an interesting avenue for future work. A sensitivity analysis of the framework with respect to numerical non-determinism in training remains an another perspective. 


\bibliography{biblio}
\end{document}